\documentclass[letterpaper, 10 pt, conference]{ieeeconf}  

\IEEEoverridecommandlockouts                              

\usepackage{color,colortbl}
\usepackage{amsmath}
\usepackage{array}
\usepackage{keystroke}
\usepackage{makecell}
\usepackage{pifont}
\usepackage{graphicx}
\usepackage{booktabs}
\usepackage{multirow}
\usepackage[export]{adjustbox}
\usepackage{balance}
\usepackage{tabularx}
\usepackage{menukeys}
\usepackage{caption}
\usepackage{fontawesome}
\usepackage{url}
\usepackage{bm}

\definecolor{materialteal}{HTML}{009688}
\definecolor{materialpurple}{HTML}{9C27B0}
\definecolor{materialindigo}{HTML}{3F51B5}
\definecolor{materialblue}{HTML}{2196F3}

\makeatletter
\let\@oldmaketitle\@maketitle
\makeatother

\title{\LARGE \bf
\sysname{}: Augmenting Robot Demonstrations through\\
Physical-World Randomization and Controlled Perturbations
}

\newcommand{\sysname}{WorldAug} 
\author{Imran Kabir$^{1}$, Yaoyu He$^{1}$, Md Alimoor Reza$^{2}$, and Syed M. Billah$^{1}$\\[3pt]
\small $^{1}$College of Information Sciences and Technology, Pennsylvania State University\\
\small $^{2}$Department of Mathematics and Computer Science, Drake University}

\newcommand{\doer}{Doer}
\newcommand{\randomizer}{Randomizer}
\newcommand{\DOne}{\mathcal{D}_1}
\newcommand{\DTwo}{\mathcal{D}_2}
\newcommand{\DThree}{\mathcal{D}_3}

\begin{document}

\maketitle
\thispagestyle{empty}
\pagestyle{empty}

\begin{abstract}
Learning from demonstration requires data that cover the visual and physical conditions a robot will encounter. Collecting this variation manually is labor-intensive, while augmentation must preserve consistency between observations and executable actions. We present \sysname{}, a system that turns a verified seed demonstration into a repeatable physical data-generation process. For each task variant, the system reuses the authored task structure: a \doer{} robot executes the task with controlled perturbations and resets the target, while a \randomizer{} robot adds and relocates physical objects between episodes. Each additional episode records observations and actions from the same physical interaction. Our prototype pairs a fixed-base Trossen WidowX arm with a Stretch~3 mobile manipulator and supports variation in target placement, distractors, illumination, and pre-grasp approaches. In a pick-and-place study with eight target objects, eight seed demonstrations produce 480 additional physical episodes. Fine-tuning $\pi_{0.5}$, a vision--language--action (VLA) model, on the generated dataset increases success from 8.33\% to 95.83\% under target-position variation and from 4.17\% to 66.67\% with 11 distractors, compared with a 16-demonstration manual baseline. MT-ACT, a policy trained from scratch, also benefits from the generated data. With varied target positions, its success rate increases from 0.00\% to 45.83\% without distractors and from 0.00\% to 29.17\% with 11 distractors.
\end{abstract}

\section{Introduction}
The performance of learned models depends strongly on the scale and diversity of their training data. In computer vision, pretraining on JFT-300M improved performance across classification, detection, and segmentation~\cite{sun2017revisiting}. In language modeling, Chinchilla demonstrated the value of increasing training tokens while balancing model size and compute~\cite{hoffmann2022compute}. Similar scaling behavior has been observed in contrastive vision--language learning~\cite{cherti2023scaling}. For robot manipulation, however, repeated demonstrations of the same scene can offer diminishing returns. Recent empirical work finds that increasing the diversity of objects and environments can matter more than increasing demonstrations within an already well-covered configuration~\cite{lin2024scaling}. This makes the collection of varied robot experience a key challenge for learning from demonstration.

Data augmentation expands training coverage by transforming existing examples. Random crops and color transformations support visual representation learning~\cite{chen2020simclr}, and back-translation produces additional parallel examples for machine translation~\cite{sennrich2016monolingual}. Robot-learning methods similarly augment observations through generative image editing, as in GenAug and ROSIE~\cite{chen2023genaug,yu2023scaling}. Yet robot demonstrations contain coupled observations and actions. Changing an object's location can change the required reach, grasp, transport path, and contact sequence. An augmentation must therefore preserve the relationship between what the robot observes and what it should do.

This requirement extends beyond successful nominal motion. A policy must distinguish the instructed object among distractors, approach it accurately, and correct deviations during execution. RoboDojo identifies weaknesses in action grounding, precision, and correction even when policies identify the appropriate task object~\cite{chen2026robodojo}. Such findings motivate demonstration coverage of off-nominal states alongside visual and spatial diversity.

Simulation offers controlled variation, repeatable resets, and parallel data generation. Domain randomization broadens simulated training conditions~\cite{tobin2017domainrandomization}, while real-to-sim-to-real pipelines such as ReBot combine simulated robot interactions with real visual context~\cite{fang2025rebot}. Their effectiveness depends on how simulated observations and control correspond to physical deployment. SIMPLER demonstrates the importance of matching these factors when evaluating real-world policies in simulation~\cite{li2024evaluating}. MimicGen and SkillMimicGen adapt demonstrated behavior to new configurations, and both include physical-robot experiments~\cite{mandlekar2023mimicgen,garrett2025skillmimicgen}. Sustained physical data collection also requires scene reconfiguration and reset between episodes. We investigate how a small seed set can support repeated physical data generation when scene variation and resetting are automated.

\begin{figure*}[t]
\centering
\includegraphics[width=0.99\linewidth]{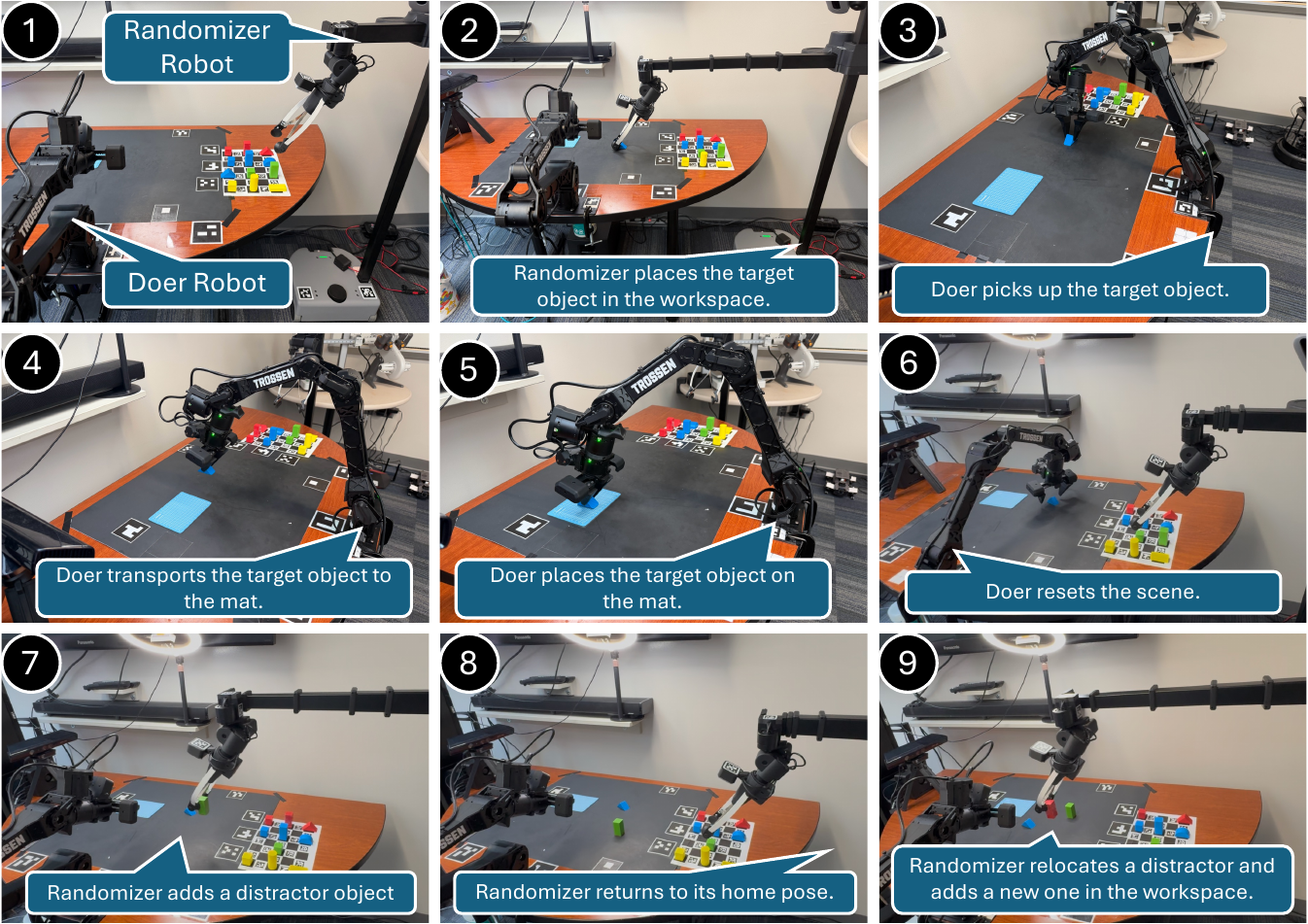}
\caption{Physical demonstration augmentation. (1--2) The \randomizer{} initializes the scene. (3--5) The \doer{} performs the pick-and-place task. (6) The \doer{} resets the target at a new location. (7--9) The \randomizer{} adds or relocates distractors between episodes. The same verified task structure seeds repeated executions under different physical configurations.}
\label{fig:teaser}
\end{figure*}

We address this question with \textbf{\textsc{\sysname{}}}, a system that augments demonstrations by changing the physical workspace and executing the resulting behavior on the target robot. A user authors and verifies one seed demonstration for each task variant. The system retains its task structure and uses it to generate additional episodes under controlled scene and execution variation. A \textit{Doer} performs the manipulation task and resets the target, and a \textit{Randomizer} adds and relocates objects between episodes (Fig.~\ref{fig:teaser}). This division of responsibilities supports heterogeneous robots. Our prototype pairs a fixed-base WidowX arm with a Stretch~3 mobile manipulator. The robots exchange task and scene information through a common framework; their motions and kinematic structures need not match.

\sysname{} integrates physical scene variation with configurable pre-grasp offsets and execution timing. Scene changes alter the robot's actual observations, obstacle geometry, and feasible motion paths. Execution perturbations introduce off-nominal approaches that can be followed by corrective motion. Each derived episode is physically executed and recorded, so its sensory observations and robot motion arise from the same interaction. Scene reconfiguration and reset prepare the workspace for the next episode. The desktop interface (Fig.~\ref{fig:sysname}) supports seed authoring and parameter configuration.

We evaluate downstream data utility on three datasets for pick-and-place with eight target objects. On each dataset, we fine-tune the pretrained vision--language--action (VLA) model $\pi_{0.5}$~\cite{team2024pi0.5} and train MT-ACT~\cite{bharadhwaj2024roboagent} from scratch. Eight seed demonstrations generate 480 additional physical episodes. Compared with a 16-demonstration manual baseline, the full generated dataset increases $\pi_{0.5}$ success by 87.50 percentage points under target-position variation without distractors and by 62.50 percentage points with all 11 distractors present. For MT-ACT, the corresponding gains in success rate are 45.83 and 29.17 percentage points. The datasets differ in size and scene coverage, so these comparisons assess the combined data-generation procedure.

Our contributions include (1) a physical-world augmentation workflow that uses a verified seed demonstration to generate additional sensorimotor episodes under scene and execution variation; (2) a heterogeneous \doer{}--\randomizer{} implementation that coordinates task execution, physical reconfiguration, and reset; and (3) a real-robot evaluation showing that the resulting data support adaptation of two policy families across target positions, colors, shapes, and clutter conditions.

\section{Related Work}
\subsection{Data for Generalist Robot Policies}
Large robot datasets illustrate the value and cost of varied physical experience. BridgeData~V2 and DROID collect demonstrations across tasks and environments to support generalization~\cite{walke2023bridgedatav2,khazatsky2024droid}. Offline imitation-learning studies also find that data quality and training choices affect how well a policy uses its demonstrations~\cite{mandlekar2022whatmatters}. These findings motivate collecting observations and actions together under the changes a policy will face at deployment.

\subsection{Interfaces for Physical Demonstration Collection}
ALOHA uses paired leader and follower arms for continuous human teleoperation~\cite{zhao2023aloha}. VR-based interfaces let an operator direct a robot without a dedicated leader arm and have supported large-scale collection in BridgeData~V2 and DROID~\cite{zhang2018deep,walke2023bridgedatav2,khazatsky2024droid}. PATO reduces the amount of continuous supervision by using a learned policy to execute familiar portions of a task and requesting human control as needed~\cite{dass2023pato}.

UMI uses handheld grippers to collect demonstrations independently of the target robot and matches sensing and actuation latency across collection and deployment~\cite{chi2024umi}. AR2-D2 captures human demonstrations through a mobile augmented-reality interface~\cite{duan2023ar2}. ARCADE uses augmented reality to collect a human demonstration and synthesizes additional trajectories by sampling and connecting waypoints~\cite{yang2024arcade}. In \sysname{}, the operator specifies the task at the object level and verifies the first physical execution. The authored specification then supports repeated executions with physical scene updates between them.

\subsection{Observation and Trajectory Augmentation}

Observation augmentation increases diversity without repeating every physical interaction. GenAug and ROSIE modify visual content while retaining associated action supervision~\cite{chen2023genaug,yu2023scaling}. DemoGen uses a single source demonstration per task to synthesize point-cloud observations and adapted action trajectories~\cite{xue2025demogen}. Real2Render2Real constructs rendered training data without object-dynamics simulation or additional robot rollouts~\cite{yu2025real2render2real}. RoboAgent also uses semantic augmentation to increase visual diversity and introduces MT-ACT for language-conditioned multi-task manipulation~\cite{bharadhwaj2024roboagent}. \sysname{} acquires a new physical observation--action trajectory for each generated episode, including the sensing and motion consequences of the changed scene.

\subsection{Automated Generation and Physical Execution}
MimicGen segments demonstrations into object-centric subtasks and transforms them to new object configurations~\cite{mandlekar2023mimicgen}. SkillMimicGen combines adapted manipulation segments with motion-planned transitions, including generation for physical manipulation tasks~\cite{garrett2025skillmimicgen}. Manipulate-Anything uses vision--language models to automate demonstration generation on real robots~\cite{duan2024manipulateanything}. These approaches reduce the need for continuous expert control, while requirements for obtaining object poses, preparing the workspace, and restoring the task state depend on the physical setup.

\sysname{} builds on this direction through a coordinated physical collection loop. It coordinates task execution, scene reconfiguration, and target reset, while retaining the verified task structure across executions. The framework exposes scene and execution variation as collection parameters.

\subsection{Simulation and Transfer}
Simulation supports parallel rollouts, controlled variation, and repeatable resets. Domain randomization varies appearance and dynamics to improve transfer~\cite{tobin2017domainrandomization}. RoboTwin~2.0 applies broad randomization to simulated manipulation data generation~\cite{chen2025robotwin}, and adaptive domain randomization uses physical observations to refine the simulation parameter distribution~\cite{chebotar2019closing}. ReBot introduces a real-to-sim-to-real video synthesis pipeline~\cite{fang2025rebot}.

The relevance of simulated experience depends on how well sensing, control, and contact reflect deployment. SIMPLER studies how matching visual and control factors affects evaluation of real-world policies in simulation~\cite{li2024evaluating}. Physical collection directly exposes the learner to the deployment robot's sensing and dynamics. Its throughput depends on hardware execution time and the ability to reset tasks reliably.

\subsection{Perturbations and Corrective Demonstrations}
Imitation policies can encounter states that are poorly represented in nominal demonstrations. DART injects noise into supervisor demonstrations to elicit correction, and includes physical grasping experiments~\cite{laskey2017dart}. IntervenGen autonomously expands a small set of corrective interventions and evaluates robustness in simulation and a physical environment~\cite{hoque2024intervengen}. \sysname{} integrates configurable execution perturbations with robot-mediated physical scene variation. Determining the incremental benefit of its perturbations requires an ablation that holds the scene distribution and data budget fixed.

\section{\sysname{}: Physical Demonstration Augmentation Framework}
\label{sec:method}

\begin{figure*}[t!]
\centering
\includegraphics[width=0.99\linewidth]{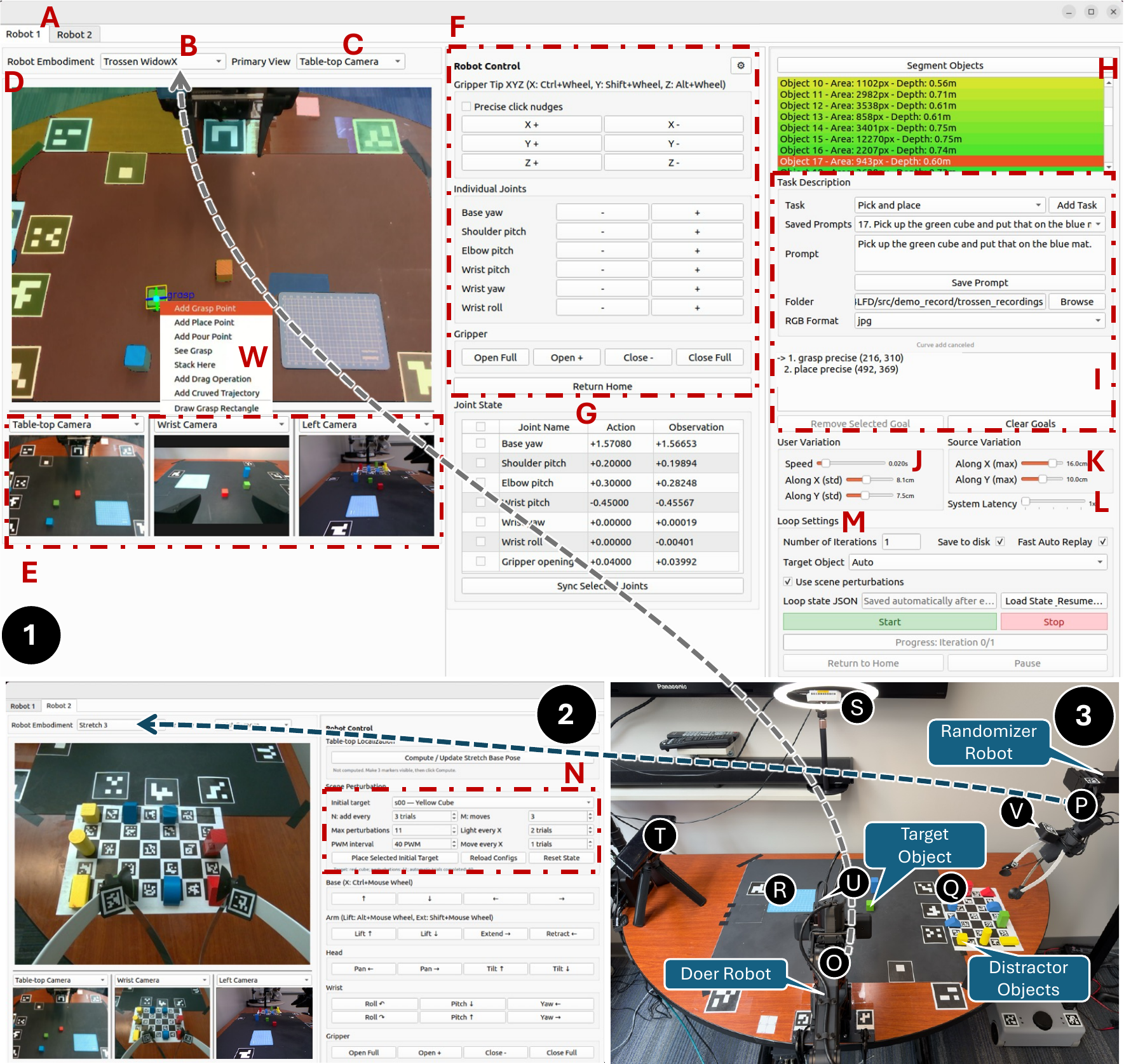}
\caption{Overview of \sysname{} and the data-collection setup. 
(1) The \doer{} interface for task authoring, robot control, and automated replay. 
(2) The \randomizer{} interface for workspace reconfiguration. 
(3) The physical setup used for data collection. 
(A) Tabs for the two robot interfaces: Robot~1 as the \doer{}, Robot~2 as the \randomizer{}.
(B--C) Robot embodiment and primary-camera selection. 
(D) Segmented primary view for selecting task objects. A secondary click on any object opens a context menu (W) containing actions and object-specific commands.
(E) Auxiliary live camera feeds. 
(F) Joint-level robot controls. 
(G) Live joint-state monitoring. 
(H) List of candidate segmented objects. 
(I) Task specification and editable action sequence. 
(J) Pre-grasp perturbation controls for inducing spatial error and recovery behavior. 
(K) Parameters for randomized object placement during replay. 
(L) System-latency control. 
(M) Automated replay and data-recording controls. 
(N) \randomizer{} settings for target selection, distractor placement and relocation, and lighting-variation intervals.
(O) The \doer{} robot (Trossen WidowX). 
(P) The \randomizer{} robot (Stretch 3). 
(Q) ChArUco-board containing the distractor objects. 
(R) Blue placement mat used as the task destination. 
(S) Tabletop Intel RealSense D435i camera with a ring light for workspace illumination. 
(T) Left-side Xbox Kinect 360 camera. 
(U-V) Intel RealSense D405 wrist camera on both robots.
}
\label{fig:sysname}
\end{figure*}

\sysname{} is a software-hardware framework for semi-automatic physical demonstration augmentation. It integrates task authoring and verification with motion planning, controlled perturbations, scene randomization, reset, and episode recording (Figs.~\ref{fig:teaser} and~\ref{fig:sysname}). It passes common grasp and placement specifications to embodiment-specific planning adapters, which generate each robot's motions.

\subsection{Task authoring and seed verification}
The framework converts a user-authored task into a verified seed for automatic collection. In Fig.~\ref{fig:sysname}, Panel A provides two tabs, \texttt{\small Robot 1} and \texttt{\small Robot 2}, to configure \doer{} and \randomizer{}, respectively. In the first tab, the user selects the embodiment and primary camera (B--C). To author ``Pick up the green cube and place it on the blue mat,'' the user selects the cube's segmentation mask (D), adds a grasp operation, and clicks the placement point on the mat. Panel I stores the instruction and editable action sequence. The user configures pre-grasp perturbations (J), target-position variation (K), and execution delay (L). Panel N provides controls for target selection and the schedules for distractor and lighting changes. The user requests additional episodes and selects \texttt{Start} (M). The framework records the first execution, pausing at the pre-grasp and pre-placement poses for verification and optional correction. This \textit{human-authored demonstration} supplies the seed. The software retains the verified grasp and placement poses, gripper commands, and relevant trajectory information for automatic replay.

The authoring interface supports point-and-click selection of segmented objects and a secondary-click menu for grasping, placement, and other commands (W). Pointer movement specifies destinations; scrolling adjusts height or orientation. Dragging specifies planar pushing, pulling, or sliding motions, while drawing specifies curved trajectories. Users compose these operations into tasks such as stacking and sorting, then inspect, reorder, or refine their steps in Panel I. Joint-level controls (F) and keyboard-modified mouse-wheel shortcuts support pose correction during seed verification, with live joint states (G) providing feedback.

\subsection{Perception and embodiment-specific execution}
The framework's perception module converts the selected target into a Cartesian grasp specification~\cite{ten2017using}. SAM~\cite{kirillov2023segment} segments the primary RGB image, and the selected mask isolates target points in the aligned depth data. We fit a plane to the object's top-surface points and project them onto the two principal axes of that plane. We compute a bounding rectangle from the projected points. Its centroid determines grasp position, its shorter side determines gripper opening, and its longer-side orientation determines grasp yaw. The gripper closes across the shorter dimension. The selected destination supplies the placement goal.

The grasp specification includes the grasp position, the end-effector orientation as a quaternion, and the gripper opening. The placement pose gives the target position and orientation. A calibrated reference frame anchored at the \doer{} base relates the tabletop camera and both robot bases. Embodiment-specific adapters receive these specifications and translate them into robot motion, using ROS for state monitoring and control and URDF models for robot geometry and kinematics.

\paragraph{\textbf{Fixed-base planning}}
For the WidowX \doer{}, the tabletop height and camera-to-base transform express the RGB-D observations in a common planning frame. Points above the tabletop provide visible object surfaces. The planner marks the space between these surfaces and the tabletop as occupied, voxelizes the result, and inflates the occupied region by a configurable clearance margin. It filters gripper points and small regions around the start and goal before searching for a path.

BIT*~\cite{gammell2015batch} plans a Cartesian end-effector position path through the resulting obstacle representation. The system simplifies and interpolates that path into waypoints, then passes them and the desired orientation to MoveIt~\cite{gorner2019moveit}. MoveIt performs inverse kinematics, checks the trajectory against the robot and collision scene, and produces joint commands subject to the configured velocity and acceleration limits.

\paragraph{\textbf{Mobile-manipulator planning}}
For the Stretch~3 \randomizer{}, whole-body inverse kinematics determines base displacement, lift, arm extension, and wrist configuration for the pre-action pose. The robot moves the end-effector to a clearance configuration before repositioning its base, then adjusts the lift and arm extension to approach the goal. The wrist camera supports marker-based final alignment when handling distractors.

\paragraph{\textbf{Grasp and placement execution}}
Both adapters move through pre-action poses before the final local approach. After reaching the grasp pose, the robot closes the gripper and raises the object to clear the tabletop before transport. At the destination it approaches from a pre-placement pose, lowers the object, releases it, and retreats. Common grasp, placement, state, and trajectory-execution interfaces let the two embodiments use the same task-level collection workflow.

\subsection{Randomization and perturbation controls}
The interface allows users to configure the spatial distribution of starting states, execution perturbations, and scene-change schedules. For target-position randomization (Fig.~\ref{fig:sysname}, K), \sysname{} samples each new starting location from a two-dimensional Gaussian on the tabletop plane. Its mean is the target's initial position in the seed demonstration. Panel K sets the maximum displacements along the X and Y axes, bounding the sampled locations separately along each axis. The \doer{} places the target at the sampled location during reset, allowing users to control the spatial coverage of subsequent demonstrations. Separate controls in Panel J set the X and Y standard deviations of pre-grasp perturbations around the nominal approach pose to elicit corrective motion during grasp execution.

Panel N controls the intervals for distractor additions, relocations, and lighting changes; selecting different physical distractors varies the scene's color and shape composition. After scene changes, BIT* replans against the updated obstacle representation, and stochastic planning can produce additional trajectory variation. Panel L sets execution delay, which remains fixed at 20~ms across all experimental datasets. The task instruction and placement destination also remain fixed for each target-object variant in our study.

\subsection{Automated physical demonstrations}
After seed verification, \sysname{} generates \textit{automated physical demonstrations}, or derived episodes, by combining the retained task specifications with newly planned paths under controlled domain randomization~\cite{tobin2017domainrandomization}. The \randomizer{} initializes the workspace for a task variant by placing the target at its starting location (Fig.~\ref{fig:teaser}). In each derived episode, the \doer{} executes the grasp, transport, placement, and return sequence without verification pauses. Once recording ends, the software commands the \doer{} to reset the target at a new initial position, then invokes any scheduled scene reconfiguration. The \randomizer{} follows the same manipulation pipeline, with object selection and placement goals supplied by the randomization procedure. To add a distractor, the framework selects an object from a known cell on the ChArUco board; to relocate one, it selects an object already in the workspace. The placement goal is an unoccupied location with clearance from the target and other objects. These grasp and placement specifications pass through the \randomizer{}'s embodiment-specific adapter, with its wrist camera supporting final grasp alignment. The robot returns home after placing the object. The software coordinates these operations and scheduled lighting changes between recorded task executions.

\subsection{Episode recording}
\sysname{} records the tabletop (Fig.~\ref{fig:sysname}, S), left-side (Fig.~\ref{fig:sysname}, T), and \doer{} wrist (Fig.~\ref{fig:sysname}, U) camera streams together with the \doer{} robot's joint states and commanded joint values at 30~Hz in our experiments. An episode begins when the \doer{} leaves its home configuration to approach the target and ends when it returns home after placement. Target reset and distractor reconfiguration occur outside this recording interval. The \randomizer{} wrist-camera stream, joint states, and commands are excluded. The task prompt, authored sequence, and replay parameters accompany each episode, pairing the physical observations and actions with the task and collection settings that produced them.

\section{Experiments}
To evaluate the effectiveness of \sysname{}, we train two policy families on human-authored and automatically generated physical robot demonstrations. We then analyze performance under target-position and distractor variation across eight target objects of different shapes and colors. We further probe the best-performing policy by systematically varying distractor number and similarity to understand which object properties it relies on.

\subsection{Experimental Setup}

The physical data-collection setup is shown in Fig.~\ref{fig:sysname}(3). We use a Trossen WidowX as the \doer{} (O) and a Stretch 3 as the \randomizer{} (P). Distractor objects are stored on a ChArUco board (Q), and the blue mat (R) serves as the placement destination. The workspace is observed using a tabletop RealSense D435i (S), a left-side Kinect 360 (T), and wrist-mounted RealSense D405 cameras on the \doer{} (U) and \randomizer{} (V). ArUco markers distributed throughout the workspace are used to compute the tabletop-camera-to-\doer{}-base transformation and the \randomizer{}-base pose relative to the \doer{} base, which serves as the global reference frame.

\subsection{Dataset}
\label{subsec:dataset}
We collected a real-world manipulation dataset using \sysname{} with a set $\mathcal{O}$ of 12 objects varying in shape and color. The target set $\mathcal{O}_\tau \subset \mathcal{O}$ contains eight objects: red, green, and blue pillars; red, green, and blue cubes; and red and blue triangular blocks. The remaining four objects are a yellow cylinder, yellow rectangular block, yellow cube, and blue arc-shaped block. For each target $o_t\in\mathcal{O}_\tau$, distractors are selected from $\mathcal{O}\setminus\{o_t\}$, allowing them to share either the target's color or shape.

For each of the eight targets, we collect one \textit{human-authored demonstration}, followed by 60 automatically generated \textit{physical robot demonstrations} using the method described in Section~\ref{sec:method}. The first 23 generated episodes vary the target's initial position without distractors. Starting at generated episode~24, the \randomizer{} adds one distractor every three episodes, reaching all 11 distractors at episode~54. 
Existing distractors are relocated every episode, provided that no new distractor is added in that episode.
After all 11 distractors have been introduced, the remaining episodes maintain this count and continue the relocation schedule. The target position continues to vary throughout the generated episodes, and lighting conditions change every two episodes.

This augmentation procedure produces eight seed demonstrations and 480 automatically generated episodes, totaling 488 demonstrations. We separately collect eight additional manual demonstrations for the baseline, bringing the full experimental collection to 496 unique demonstrations. We construct three training datasets:

\textbf{$\boldsymbol{\DOne{}}$:} The eight initial \textit{human-authored demonstrations} without distractors and eight additional \textit{human-authored demonstrations}, one for each target with all 11 distractors present. In total, $\DOne{}$ contains 16 manual demonstrations.

\textbf{$\boldsymbol{\DTwo{}}$:} 184 automatically generated demonstrations, corresponding to the first 23 generated episodes for each target, with target-position variation and no distractors.

\textbf{$\boldsymbol{\DThree{}}$:} 480 automatically generated demonstrations with both target-position and distractor variation. $\DTwo{}$ is a subset of $\DThree{}$. All manual demonstrations in $\DOne{}$ are excluded from both automatically generated datasets.

\textbf{\textit{Minimizing Human Effort.}}
Our semi-automatic collection approach substantially reduces the human time required to record demonstrations. Recording a single manual pick-and-place demonstration with one target and no distractors takes approximately one minute, including task authoring and scene reset. At this rate, manually collecting the 488 demonstrations in the augmentation run would require approximately 488 minutes of human effort, with potentially additional time needed to add, reposition, and reset distractors between episodes.

With \sysname{}, users author one initial demonstration for each target object, requiring approximately eight minutes for the eight targets, after which the remaining 480 episodes are generated automatically. Under the one-minute-per-demonstration assumption, this corresponds to an estimated reduction of approximately 98\% in human demonstration-collection time. This estimate compares seed authoring with repeated manual recording for the augmentation run; the eight additional manual baseline demonstrations are separate from this calculation.

\subsection{Baselines}
We evaluate two policy families. $\pi_{0.5}$~\cite{team2024pi0.5} is a pretrained VLA policy with strong reported performance on RoboDojo's real-world benchmark~\cite{chen2026robodojo}. Fine-tuning it tests whether generated demonstrations support adaptation of a generalist policy to our task and workspace. MT-ACT~\cite{bharadhwaj2024roboagent} is a language-conditioned multi-task transformer that predicts chunks of actions. Training it from scratch tests whether the benefits extend to a policy trained on the task demonstrations alone. For each family, we train a separate policy on $\DOne{}$, $\DTwo{}$, and $\DThree{}$ using the same architecture and training schedule across datasets.

\subsection{Implementation}
All demonstrations are recorded at 30~Hz, and the configured execution delay is fixed at 20~ms across datasets.

\subsubsection{\textbf{$\pi_{0.5}$}}
We use Physical Intelligence's official \texttt{openpi} fine-tuning script~\cite{githubOpenpiREADMEmdMain} and initialize each policy from the released \texttt{pi05\_base} checkpoint. We fine-tune for 80,000 iterations on four NVIDIA RTX A6000 GPUs with 48~GB of memory each and an action chunk size of 30. The remaining training parameters follow the official implementation.

\subsubsection{\textbf{MT-ACT}}
We use the official MT-ACT implementation from the RoboAgent repository~\cite{githubRoboAgent} and train for 30,000 epochs on one NVIDIA RTX A6000 GPU with an action chunk size of 60. The remaining training parameters follow the official implementation.

\subsection{Evaluation Protocol}
\label{subsec:eval_protocol}
We evaluate the trained policies under controlled variations in the target-object position and distractor configuration. The evaluation uses the same target identities as the training datasets.

\subsubsection{\textbf{No-distractor condition}}
We first evaluate each policy with only the target object. Let $p_t$ denote the target position in an evaluation trial, $p_0$ a target position observed in $\DOne{}$, and $\mathcal{W}$ the valid workspace. We evaluate both policy families with target positions varied across the workspace ($p_t \sim \mathcal{W}$). We also evaluate $\pi_{0.5}$ with approximately fixed target positions ($p_t \approx p_0$). We run three rollouts per target, for 24 rollouts per policy and setting across the eight targets.

\subsubsection{\textbf{Full-distractor condition}}
We also evaluate each target object with all 11 remaining objects present as distractors. The target position is varied across the workspace, $p_t \sim \mathcal{W}$, while distractor positions are also varied between trials. We evaluate both policy families trained on $\DOne{}$ and $\DThree{}$ in this condition. This represents the most cluttered configuration supported by our object set $\mathcal{O}$. Similar to the no-distractor condition, we run three evaluation rollouts for each of the eight target objects.

\subsubsection{\textbf{Systematic distractor variation}}
Between these two boundary conditions, we systematically vary distractor similarity and number for the best-performing policy, $\pi_{0.5}$ trained on $\DThree{}$. We use $c$ and $s$ to denote object color and shape, respectively. Subscripts $t$ and $d$ denote the target and distractors. The number of distractors is varied as $N_d \in \{1,2,4,6\}$, and distractor positions are varied between trials. We evaluate four conditions:

\begin{itemize}
    \item[\textbf{I:}] $c_t\neq c_d, s_t=s_d, p_t\approx p_0$: same-shape, different-color distractors, with approximately fixed target position.

    \item[\textbf{II:}] $c_t=c_d, s_t\neq s_d, p_t\approx p_0$: same-color, different-shape distractors, with approximately fixed target position.
    
    \item[\textbf{III:}] $c_t\neq c_d, s_t=s_d, p_t\sim\mathcal{W}$: same-shape, different-color distractors, with the target position varied across the workspace.

    \item[\textbf{IV:}] $c_t=c_d, s_t\neq s_d, p_t\sim\mathcal{W}$: same-color, different-shape distractors, with the target position varied across the workspace.

\end{itemize}

The systematic distractor sweep comprises 432 rollouts.

\subsubsection{\textbf{Evaluation metrics}}
We report task success rate ($\mathrm{SR}$) and task progress ($\mathrm{Prog.}$). $\mathrm{SR}$ denotes the fraction of rollouts that successfully complete the full pick-and-place task, whereas $\mathrm{Prog.}$ quantifies partial task completion across five sequential stages: reaching the target, grasping the target, reaching the destination, placing the object, and returning the robot to its initial configuration.

\begin{table}[t]
\centering
\begin{tabular}{l|cc|cc}
\toprule
&
\multicolumn{2}{c|}{\textbf{Fixed Target Position}} &
\multicolumn{2}{c}{\textbf{Varied Target Position}} \\
\cmidrule(lr){2-3}
\cmidrule(lr){4-5}
\textbf{Training Data} &
\textbf{Prog. ($\uparrow$)} &
\textbf{SR ($\uparrow$)} &
\textbf{Prog. ($\uparrow$)} &
\textbf{SR ($\uparrow$)} \\
\midrule
$\mathcal{D}_1$ &
41.67\% & 25.00\% &
11.67\% & 8.33\% \\

$\mathcal{D}_2$ &
63.33\% & 54.17\% &
57.50\% & 45.83\% \\

\textbf{$\mathcal{D}_3$} &
\textbf{100.00}\% & \textbf{100.00}\% &
\textbf{96.67}\% & \textbf{95.83}\% \\
\bottomrule
\end{tabular}
\caption{Performance of $\pi_{0.5}$ under fixed and varied target positions without distractors. Each policy is evaluated over 24 trials per condition.}
\label{tab:target_position}
\end{table}

\begin{table}[t]
\centering
\begin{tabular}{l|cc}
\toprule
\textbf{Training Data} &
\textbf{Prog.  ($\uparrow$)} &
\textbf{SR  ($\uparrow$)} \\
\midrule
$\DOne{}$ &
11.67\% & 4.17\% \\

\textbf{$\DThree{}$} &
\textbf{68.33}\% & \textbf{66.67}\% \\
\bottomrule
\end{tabular}
\caption{Performance of $\pi_{0.5}$ with varied target positions and all 11 non-target objects present as distractors. Each policy is evaluated over 24 rollouts.}
\label{tab:all_distractors}
\end{table}

\begin{table}[t]
  \centering
  \begin{tabular}{l|cc|cc}
    \toprule
    & \multicolumn{2}{c|}{\textbf{No Distractors}} & \multicolumn{2}{c}{\textbf{11 Distractors}} \\
    \cmidrule(lr){2-3} \cmidrule(lr){4-5}
    \textbf{Training Data} & \textbf{Prog. ($\uparrow$)} & \textbf{SR ($\uparrow$)} & \textbf{Prog. ($\uparrow$)} & \textbf{SR ($\uparrow$)} \\
    \midrule
    $\DOne{}$ & 0.00\%  & 0.00\%  & 0.00\%  & 0.00\%  \\
    $\DTwo{}$ & 35.00\% & 33.33\% & --      & --      \\
    $\DThree{}$ & \textbf{50.00}\% & \textbf{45.83}\% & \textbf{29.17}\% & \textbf{29.17}\% \\
    \bottomrule
  \end{tabular}
  \caption{Performance of MT-ACT with varied target positions, without distractors and with all 11 non-target objects present as distractors. Each evaluated setting uses 24 rollouts. Dashes indicate unevaluated settings. Because all demonstrations in $\DTwo{}$ are distractor-free, we evaluated the policy trained on $\DTwo{}$ only in the no-distractor condition.}
  \label{tab:mtact}
\end{table}

\begin{figure}[t]
\centering
\includegraphics[width=0.99\linewidth]{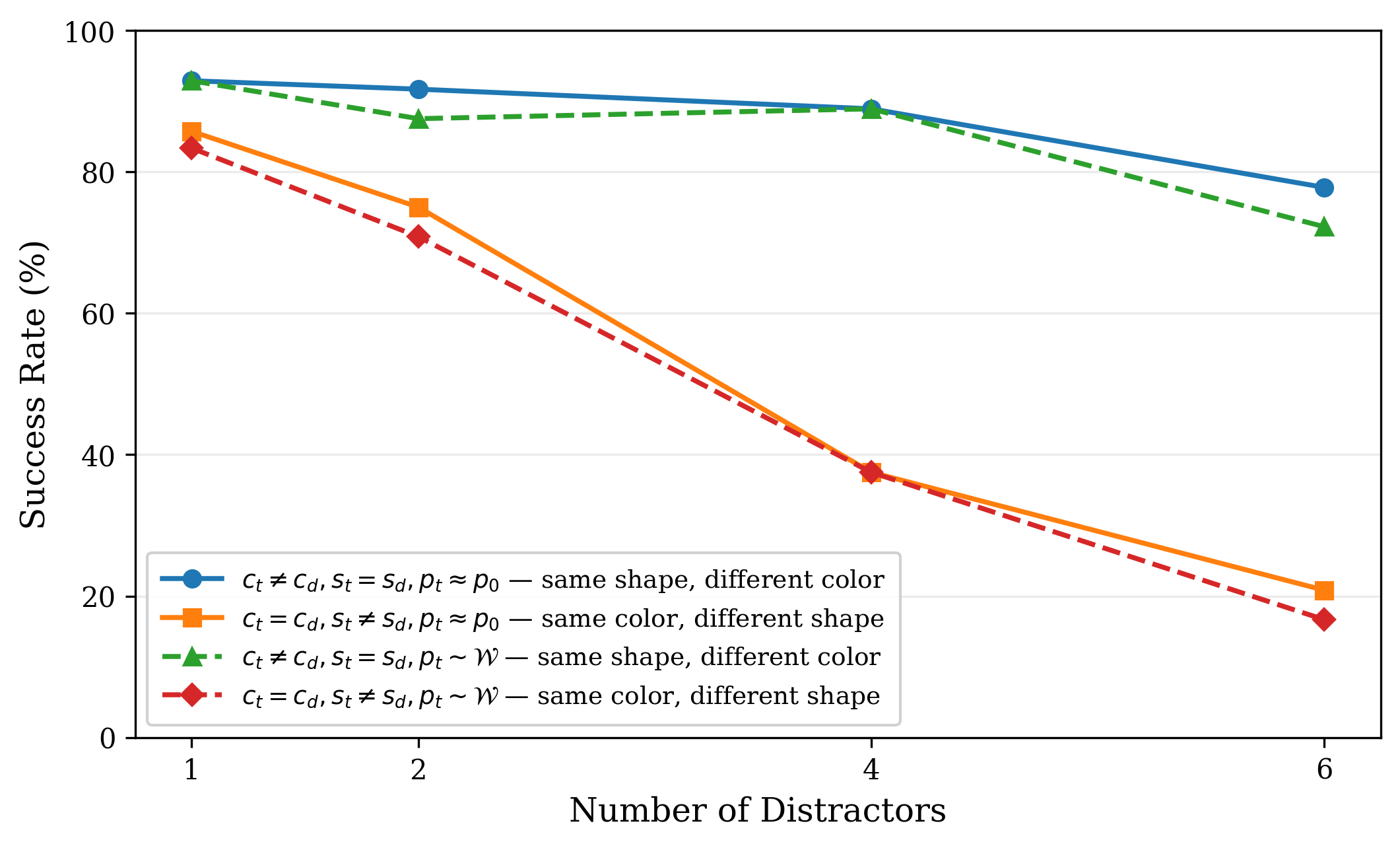}
\caption{Success rate ($SR$) under systematic distractor variation across Conditions (I)-(IV) (see Section~\ref{subsec:eval_protocol}) as the number of distractors increases. The lower pair of line plots (\textit{in orange-red}) shows that success rate degrades substantially as the number of distractors sharing the target color increases. In contrast, the top pair of line plots (\textit{in blue-green}) shows that performance remains robust in the presence of same-shape, different-color distractors (best viewed in color).}
\label{fig:distractor_var}
\end{figure}

\subsection{Results}

We report results on overall policy performance, qualitative execution behavior and partial-task failures, and the influence of distractor similarity on policy performance.

\subsubsection{\textbf{Does automatically generated data improve policy performance}}
Table~\ref{tab:target_position} compares $\pi_{0.5}$ trained on $\DOne{}$, $\DTwo{}$, and $\DThree{}$ under the no-distractor condition for both fixed and varied target positions. 

With the target position held fixed across rollouts, $\pi_{0.5}$ trained on $\DOne{}$ achieves a success rate of 25.00\% and task progress of 41.67\%. Training on $\DTwo{}$ improves performance to 54.17\% success and 63.33\% task progress. When trained on $\DThree{}$, $\pi_{0.5}$ achieves a 100\% success rate.

When the target position is varied across the workspace, the success rate decreases to 8.33\% for $\DOne{}$ and 45.83\% for $\DTwo{}$. In contrast, $\DThree{}$ maintains a success rate of 95.83\%, corresponding to only a 4.17 percentage-point reduction relative to the fixed-target condition. These results suggest that the data generated through physical \textit{domain randomization} improve the policy's ability to generalize across target locations.

With varied target positions and no distractors, MT-ACT achieves 0.00\%, 33.33\%, and 45.83\% success on $\DOne{}$, $\DTwo{}$, and $\DThree{}$, respectively (Table~\ref{tab:mtact}).

The gains also extend to the maximum-distractor condition. With all 11 non-target objects present in the workspace, $\pi_{0.5}$ trained on $\DOne{}$ achieves an $SR$ of 4.17\%, compared with 66.67\% for $\DThree{}$ (Table~\ref{tab:all_distractors}). For MT-ACT, success rises from 0.00\% on $\DOne{}$ to 29.17\% on $\DThree{}$ (Table~\ref{tab:mtact}).

Overall, automatically generated data yields substantial performance gains over the manually collected baseline while allowing additional training demonstrations to be collected without repeated human authoring. Because dataset size and variation increase together, these comparisons reflect their combined effects rather than isolating the contribution of each type of variation.

\subsubsection{\textbf{How does policy behavior differ across training datasets}}
Qualitative inspection reveals that policies trained on both $\DOne{}$ and $\DTwo{}$ can reach the target but may fail during grasping or placement when the test-time wrist-camera observations or gripper-object alignment differ from those encountered during training. This failure mode is more pronounced for $\DOne{}$, which contains a limited number of demonstrations.
These partial-task failures are reflected in the gap between task progress $Prog.$ and $SR$, which is 16.67 percentage points ($41.67-25.00$) for $\DOne{}$ in the \textit{fixed-target-position} setting and 11.67 percentage points ($57.50-45.83$) for $\DTwo{}$ in the \textit{varied-target-position} setting (Table~\ref{tab:target_position}).

Qualitatively, we observe that $\pi_{0.5}$ trained on $\DThree{}$ recovers more reliably from small positioning errors during grasping and is more consistently able to complete both grasping and placement. The difference between its task progress, $Prog.$ (96.67\%), and success rate, $SR$ (95.83\%), is only 0.84 percentage points under varied target positions. Together with the qualitative observations, the high success rate indicates more consistent completion of the grasp-and-place sequence.

\subsubsection{\textbf{How does distractor similarity affect policy performance}}
Fig.~\ref{fig:distractor_var} shows that the success rate degrades substantially as the number of distractors sharing the target color increases. With the target position approximately fixed, the $SR$ in Condition~II (same-color/different-shape distractors) decreases from 85.71\% with one distractor to 20.83\% with six distractors, whereas Condition~I (same-shape/different-color distractors) decreases only from 92.86\% to 77.78\%. A similar trend is observed under workspace-wide \textit{target position variation}. With six distractors, Condition~IV (same-color/different-shape distractors) achieves an $SR$ of 16.67\%, compared with 72.22\% for Condition~III (same-shape/different-color distractors). These results suggest that \textit{\textbf{the learned policy relies more strongly on color than on shape}} when discriminating the target from visually similar distractors. 

\textit{Target position variation} introduces comparatively little additional performance degradation. Across all distractor counts, the difference between Conditions~I and III, and between II and IV, is at most 5.56 percentage points. This suggests that the policy is relatively robust to variations in target position, consistent with the broad target-position variation introduced during \sysname{} data generation.

\section{Discussion and Future Work}
\subsection{What the dataset comparison measures}
All policies within a model family use the same training schedule. The datasets differ in the amount of experience and in coverage of target positions, appearance, clutter, and the trajectories produced by replanning. The gains therefore measure the utility of the combined collection workflow. A size-matched comparison could vary one collection parameter at a time to distinguish the effects of scene diversity, trajectory variation, and induced correction. The estimate of human time compares seed authoring with repeated manual recording under the stated assumptions. End-to-end throughput also depends on robot execution and scene reset time.

\subsection{Hardware and workspace requirements}
The current grasp estimator uses a visible top surface and a geometry suitable for a parallel-jaw gripper. Irregular or occluded objects may require a different grasp estimator. Deploying the framework on another embodiment requires compatible state and control interfaces, a robot-specific planning adapter, and calibration into the shared workspace frame. Automatic scene updates in this study also rely on a registered distractor source and objects the \randomizer{} can grasp and place reliably.

\subsection{Failure-guided randomization} The policy's greater sensitivity to same-color distractors suggests that augmentation should target specific perceptual ambiguities. Future collection could use policy failures to prioritize scenes with same-color, different-shape distractors, while controlling clutter count and layout. An adaptive collection schedule could interleave these targeted scenes with broad randomization, retaining coverage beyond previously observed errors. Evaluation on held-out layouts under a fixed collection budget would test if failure-guided sampling improves generalization more efficiently than the current predefined schedule.

\subsection{Extending task coverage}
We evaluate pick-and-place for known target objects under varying positions and scene clutter. The framework separates task authoring from embodiment-specific execution, providing a basis for extending both the demonstration interface and the supported manipulation tasks. Virtual-reality input and passthrough-camera views could support freehand seed authoring beyond the current point-and-click interaction. Applying the framework to folding laundry, pouring liquids, or tasks involving changes in grasp would also require suitable hardware, sensing, and task controllers. A seed captures one manipulation strategy, so additional seeds may be needed when scene changes require different grasps or contact sequences. Scene reconfiguration and reset would likewise depend on the task, including restoring a garment configuration or replenishing a liquid container. These extensions could retain the coordination between task execution and scene preparation, while adapting how each operation is performed. Evaluating them would establish how well the collection approach extends beyond the reference task studied here.

\subsection{Scaling physical collection} The architecture could extend to one mobile \randomizer{} serving several \doer{} workspaces, preparing one while another records an episode. Scaling would then depend on whether travel and reconfiguration keep pace with task execution, together with coordinated workspace access during handoffs. Accepted episodes per robot-hour and active human time per accepted episode would quantify collection efficiency as robot count increases, accounting for reset failures and interventions. These measures help determine when additional robots justify their hardware and coordination costs.

\section{Conclusion}
\sysname{} augments robot demonstrations through coordinated physical task execution, scene randomization, controlled perturbations, and reset. A verified seed supplies reusable task structure, while heterogeneous \doer{} and \randomizer{} robots generate additional episodes under varied physical conditions. Our pick-and-place study shows that fine-tuning $\pi_{0.5}$ on these episodes improves task success across target positions and clutter conditions compared with fine-tuning on a small manually collected dataset. Training MT-ACT from scratch on these episodes also improves success under varied target positions without distractors and with all 11 distractors present. These findings support physical-world augmentation as a promising approach to expanding demonstration coverage, with minimal human intervention after seed verification.

\bibliographystyle{IEEEtran}
\bibliography{local}

\end{document}